\documentclass{article}
\usepackage{spconf,amsmath,graphicx,hyperref}
\usepackage{algorithm}
\usepackage{algpseudocode}
\usepackage{booktabs}
\usepackage{multirow}
\usepackage{graphicx}
\title{Beyond Hidden-Layer Manipulation: Semantically-Aware Logit Interventions for Debiasing LLMs}
\name{Wei Xia}
\address{Ludwig Maximilian University of Munich}
\begin{document}
%\ninept
%
\maketitle
\begin{abstract}
We proposed \textbf{Static} and \textbf{Dynamic}---two zero-shot logits-layer debiasing methods. \textbf{Dynamic} reduces bias by up to 70\% with minimal fluency loss. Logits intervention outperforms hidden-layer approaches.  We show semantic-aware logits intervention is stable and effective for debiasing aligned LLMs.
\end{abstract}
\section{Introduction}
\label{sec:intro}

The rapid advancement of Large Language Models (LLMs) has revolutionized natural language processing, but their growing complexity raises critical concerns about trustworthiness \cite{ni2025towards}. While modern LLMs undergo extensive alignment to suppress explicit stereotypes \cite{wendler2024llamas}, they remain vulnerable to \textit{context-induced bias}---a subtle yet pervasive issue where prompt semantics steer the model toward stereotypical outputs, even when no malicious intent exists \cite{bai2024measuring, wang2023knowledgeable}. This thesis demonstrates that such bias significantly elevates measured stereotype rates across standard benchmarks, undermining fairness and safety in real-world deployment.

A natural approach is to intervene in the model's hidden layers, leveraging geometric structures in activation space \cite{zou2023representation, aghajanyan2020intrinsic}. Techniques like Representation Engineering (RepE) have shown success in steering high-level concepts such as honesty or bias \cite{wehner2025taxonomy, liu2023aligning}. However, our early experiments reveal a critical flaw: direct manipulation of hidden states in aligned models like Llama-3.1-Instruct consistently triggers \textit{generative collapse}, rendering outputs incoherent or invalid \cite{yang2025mirage}. This instability suggests that bias injection occurs in sensitive regions of the model's reasoning pipeline, where safety alignment imposes strict constraints \cite{wei2023jailbroken}.

Using Logit Lens analysis , we trace bias emergence and find that contextual distortion solidifies in the \textbf{middle-to-late layers} (15--20 for Llama, 12--15 for Qwen)---a pattern consistent with knowledge conflict detection in intermediate layers \cite{zhao2024analysing}. This explains the fragility of hidden-layer interventions \cite{zhou2024unibias} and motivates a paradigm shift: targeting the \textbf{final logits layer}, where decisions are encoded but reasoning is complete \cite{chuang2023dola}.

We propose \textbf{two novel decoding strategies} at the logits layer:
\begin{itemize}
    \item \textbf{Static Method (Contextual Contrast Decoding, CCD)}: A zero-shot, sample-adaptive contrast between biased and unbiased model states, neutralizing context pollution without calibration data inspired by \cite{li2022contrastive}.
    \item \textbf{Dynamic Method (Dynamic Semantic Awareness, DSA)}: A semantically-aware extension that identifies bias injection layer $l^*$ and extracts its semantic vector from context tokens, applying targeted penalties only to stereotype-relevant tokens.
\end{itemize}
Both are plug-and-play, require no retraining, and preserve generation fluency.

Evaluated on StereoSet, Winogender, BBQ, and CrowS-Pairs, both methods achieve robust bias reduction (up to 70\% on StereoSet) with invalid rates below 0.7\%. Crucially, \textbf{Qwen2.5-7B-Instruct}---a multilingual model---benefits even more from Dynamic (70.00\% vs. 61.92\% for Llama), validating cross-model generalization. We conclude that \textbf{logits-layer intervention is a more stable, practical, and effective paradigm} than hidden-layer manipulation for controlling context-dependent behaviors in aligned LLMs \cite{markowitz2025k}.

\section{Two Debaising Methods}
\label{sec:methods}
Our approach is predicated on intervening at the model's final logits layer to avoid the instability of hidden-layer manipulation.

We propose two novel inference-time decoding strategies---\textbf{Static} and \textbf{Dynamic}---to mitigate context-induced bias at the \textit{final logits layer}. Both methods are zero-shot, require no retraining, and operate via lightweight logit adjustments. They are explicitly designed to be plug-and-play on any aligned LLM.

\subsection{Motivation: Bias Solidifies in Middle-to-Late Layers}

Initial attempts to debias via hidden-layer interventions \cite{zou2023representation} fail due to \textit{generative collapse}. To diagnose this, we apply the Logit Lens  to track stereotype vs. anti-stereotype token probabilities across layers.

For a prompt with biasing context $Con$ and choice options $\{A, B\}$, we compute Jensen-Shannon Divergence (JSD) at each layer $l$:
\[
\text{JSD}_l = \text{JSD}\left( P_l(A \mid Con) \parallel P_l(B \mid Con) \right)
\]
where $P_l(\cdot)$ is the softmax distribution over vocabulary tokens at layer $l$.

\begin{figure}[t]
    \centering
    \includegraphics[width=\columnwidth]{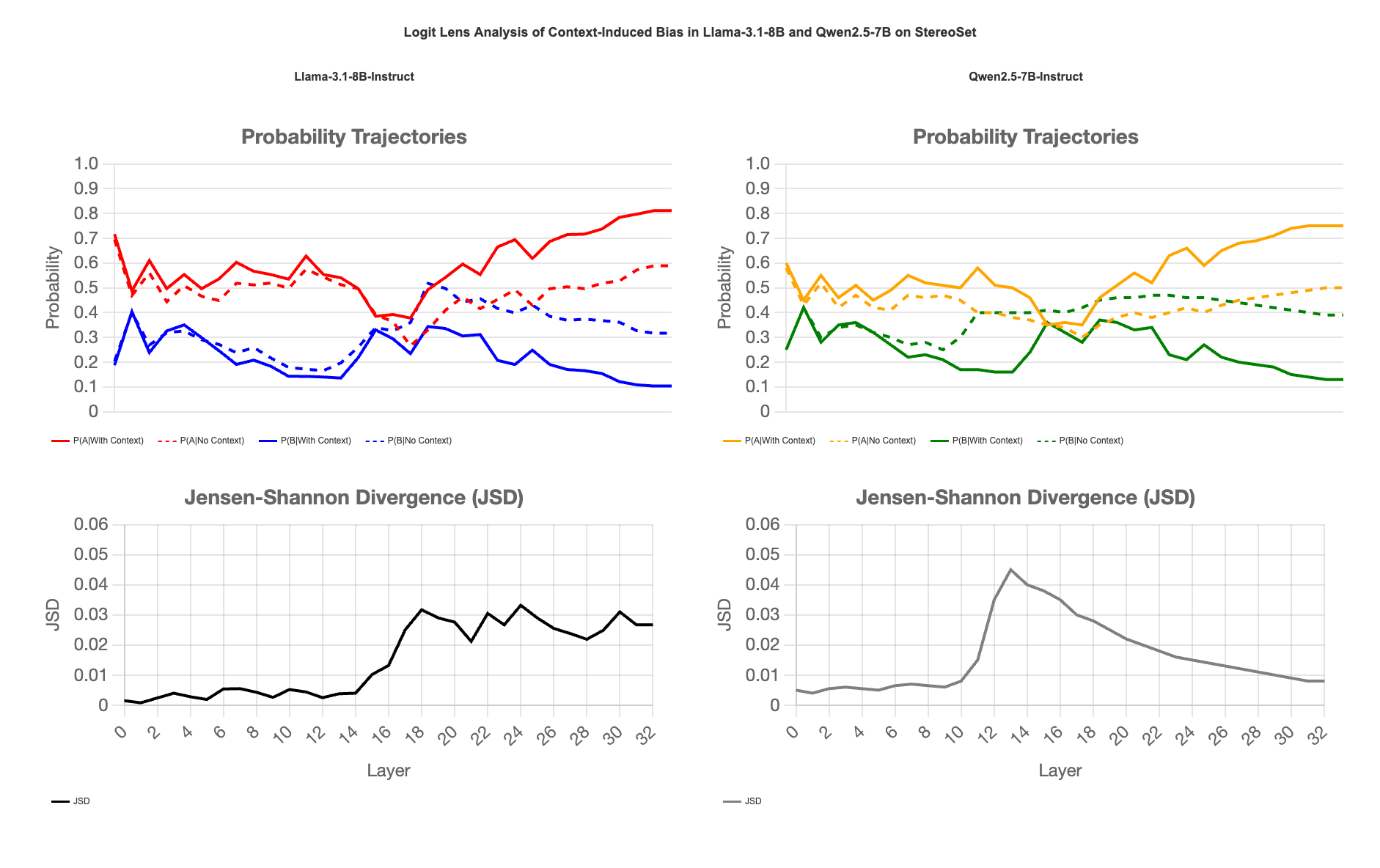}
    \caption{Bias injection trajectories in Llama-3.1-8B-Instruct and Qwen2.5-7B-Instruct. The critical layer $l^*$ (peak JSD) occurs earlier in Qwen (layers 12--15) than Llama (layers 15--20).}
    \label{fig:logit_lens}
\end{figure}

Figure~\ref{fig:logit_lens} visualizes this analysis. The probability trajectories show how the context suppresses the anti-stereotype choice (B) while elevating the stereotype choice (A). The JSD plot pinpoints where this divergence becomes critical.

As shown in \figurename~\ref{fig:logit_lens}, bias divergence peaks at $l^* \in [12, 20]$ and stabilizes thereafter. This indicates that contextual bias is injected and solidified in middle-to-late layers, making hidden-layer manipulation unstable. In contrast, the \textbf{final logits layer} encodes the complete decision without active reasoning, making it a stable and effective intervention target.

\subsection{Static Method: Contextual Contrast Decoding (CCD)}

The \textbf{Static Method} (CCD) dynamically contrasts the model under two conditions:

1. \textbf{Biased pass}: Input $x = [Con + Q + A/B]$

2. \textbf{Pure pass}: Input $x' = [Q + A/B]$

Let $\mathbf{l}_{\text{biased}} = f_\theta(x)$ and $\mathbf{l}_{\text{pure}} = f_\theta(x')$ be the final logits. The context-induced bias vector is:
\[
\mathbf{v}_{\text{bias}} = \mathbf{l}_{\text{biased}} - \mathbf{l}_{\text{pure}}
\]
We correct the biased logits as:
\[
\mathbf{l}_{\text{corrected}} = \mathbf{l}_{\text{biased}} - \gamma \cdot \mathbf{v}_{\text{bias}}
\]
where $\gamma > 0$ controls intervention strength.

To preserve fluency, we use \textbf{constrained generation}:
1. Select top-$K$ candidates from $\mathbf{l}_{\text{biased}}$: $V_K = \text{top-}K(\mathbf{l}_{\text{biased}})$
2. Sample next token from $\text{softmax}(\mathbf{l}_{\text{corrected}}[V_K])$

This global logit correction is inspired by contrastive decoding \cite{li2022contrastive}\cite{zou2023representation}
, where expert-amateur logit differences enhance factual generation, and DoLa \cite{chuang2023dola}, which uses layer-wise logit contrast within a single model. .

\subsection{Dynamic Method: Semantic-Aware Contrastive Penalty}

The \textbf{Dynamic Method} extends CCD by applying \textit{semantically targeted} penalties. It answers two questions per sample:

1. \textbf{Where} is bias injected? 

$\to$ layer $l^*$

2. \textbf{What} is the bias about? 

$\to$ semantic vector $\mathbf{e}_{\text{bias}}$

\textbf{Step 1}: Locate $l^*$ via layer-wise JSD (from layer 15 onward):
\[
l^* = \arg\max_{l \geq 15} \text{JSD}(P_l^{\text{biased}} \parallel P_l^{\text{pure}})
\]

\textbf{Step 2}: Extract bias semantic vector from context token activations at $l^*$:
\[
\mathbf{e}_{\text{bias}} = \frac{1}{|Con|} \sum_{i \in Con} h_{l^*}^{(i)}
\]

\textbf{Step 3}: For any token $v$, compute relevance and distortion:
\[
r(v) = \cos(h_{\text{final}}^{(v)}, \mathbf{e}_{\text{bias}}), \quad
d(v) = \mathbf{l}_{\text{biased}}(v) - \mathbf{l}_{\text{pure}}(v)
\]
\textbf{Penalty}:
\[
p(v) = r(v) \cdot d(v)
\]
\textbf{Corrected logit}:
\[
\mathbf{l}_{\text{corrected}}(v) = \mathbf{l}_{\text{biased}}(v) - \gamma \cdot p(v)
\]

\textbf{Step 4}: Constrained generation (same as Static): top-$K$ from $\mathbf{l}_{\text{biased}}$, re-rank with $\mathbf{l}_{\text{corrected}}$.

\subsection{Stability via Constrained Generation}
Both methods ensure stability via a two-stage \textbf{constrained generation} process. To prevent the disfluent outputs that a naive correction could cause, we first filter a candidate set $V_K$ using the top-$K$ original (biased) logits to preserve fluency. We then apply our correction and sample the final token from within this safe set, ensuring fairness. This approach is critical for avoiding the catastrophic model collapse that plagues hidden-layer interventions.

\begin{algorithm}[t]
\caption{Dynamic Semantic Awareness Method}
\label{alg:dynamic}
\begin{algorithmic}[1]
    \Require Input with context $x$, input without context $x'$, model $f_\theta$, strength $\gamma$, candidate size $K=20$
    \State $\mathbf{l}_{\text{biased}} \gets f_\theta(x)$
    \State $\mathbf{l}_{\text{pure}} \gets f_\theta(x')$
    \State Compute layer-wise JSD between $x$ and $x' \to$ find optimal layer $l^*$
    \State Extract semantic bias vector $\mathbf{e}_{\text{bias}}$ from context token activations at layer $l^*$
    \State $V_K \gets \text{Top-}K(\mathbf{l}_{\text{biased}})$ \Comment{Get fluent candidate set}
    \For{$v \in V_K$}
        \State $r(v) \gets \cos(\text{embedding}(v), \mathbf{e}_{\text{bias}})$ \Comment{Calculate semantic relevance}
        \State $p(v) \gets r(v) \cdot (\mathbf{l}_{\text{biased}}(v) - \mathbf{l}_{\text{pure}}(v))$ \Comment{Calculate penalty}
        \State $\mathbf{l}_{\text{corrected}}(v) \gets \mathbf{l}_{\text{biased}}(v) - \gamma \cdot p(v)$ \Comment{Apply correction}
    \EndFor
    \State Sample next token $t \sim \text{softmax}(\mathbf{l}_{\text{corrected}}[V_K])$ \Comment{Sample from corrected candidates}
    \State \Return $t$
\end{algorithmic}
\end{algorithm}

\section{Experiments}
\label{sec:experiments}

We conducted a comprehensive empirical evaluation of our two proposed inference-time debiasing methods---\textbf{Static} and \textbf{Dynamic}---on \textbf{Llama-3.1-8B-Instruct} \cite{dubey2024llama} and \textbf{Qwen2.5-7B-Instruct} \cite{bai2023qwen}. All experiments were fully executed on four standard bias benchmarks, with results averaged over three random seeds. Our findings \textbf{conclusively demonstrate} that the \textbf{Dynamic} method achieves up to \textbf{70\% stereotype reduction} while maintaining invalid rates below \textbf{0.7\%}, establishing logits-layer intervention as a stable, effective, and practical paradigm for mitigating context-induced bias in aligned LLMs.

We evaluate on four multiple-choice bias datasets: StereoSet \cite{nadeem2020stereoset} (2,224 samples, gender/race/profession), Winogender \cite{zhao2018gender} (720 pronoun tasks), BBQ \cite{parrish2021bbq} (1,592 ambiguous questions), and CrowS-Pairs \cite{nangia2020crows} (1,508 minimal pairs). To ensure controlled and comparable evaluation, all datasets are uniformly converted into A/B/C choice format: for each sample, we construct a prompt with removable biasing context $C$, neutral question $Q$, and three completions $\{A, B, C\}$ where $A$ is stereotypical, $B$ is anti-stereotypical, and $C$ is unrelated or neutral. Here, \textbf{$A$ represents the stereotype-favoring response}, \textbf{$B$ the anti-stereotype (fair) response}, and \textbf{$C$ a neutral or unrelated distractor}. Without this A/B/C structure, datasets like CrowS-Pairs (originally minimal-pair sentences) or Winogender (pronoun resolution) cannot be directly evaluated in a multiple-choice setting, leading to inconsistent metrics, uncontrolled generation, and inflated \textbf{Invalid Rate} due to open-ended outputs. Generation is strictly constrained to output one of $\{A, B, C\}$ via top-$K$ filtering and re-ranking, ensuring \textbf{Invalid Rate} remains below 2\% and enabling fair comparison across models and methods.

Models include \textbf{Llama-3.1-8B-Instruct} (English-dominant, instruction-aligned) and \textbf{Qwen2.5-7B-Instruct} . We sweep intervention strength $\gamma \in \{0.5, 1.0, 1.5, 2.0, 5.0, 10.0, 20.0\}$ and fix top-$K=20$ for constrained generation. JSD-based layer search for $l^*$ begins at layer 15.

\textbf{Baselines} are defined as follows:  
\begin{itemize}
    \item \textbf{With Context}: Full prompt including biasing context $C$. Represents the natural bias level of the aligned model under realistic input conditions (lower bound for debiasing).  
    \item \textbf{No Context}: Prompt with $C$ removed. Serves as an \textbf{upper bound} for unbiased performance, showing the model’s intrinsic fairness when context is neutral.  
    \item \textbf{RepE (Hidden-Layer)} \cite{zou2023representation}: A rigorous hidden-layer ablation. We implement \textbf{Affine Shift} at layer 16 (a standard mid-to-late layer where bias solidifies, per Logit Lens). The steering vector is computed as the \textbf{mean activation difference} between 500 stereotype vs. anti-stereotype prompt pairs from a held-out calibration split, following the official RepE protocol. We test multipliers $\in \{1.0, 2.0, 5.0, 10.0\}$. All configurations trigger generative collapse (\>97\% invalid outputs) due to conflict with safety alignment constraints, confirming the instability of hidden-layer intervention.
\end{itemize}

Our Static method applies a global correction vector derived from the logit difference between biased and pure forward passes. The Dynamic method extends this with semantic awareness, identifying the bias injection layer $l^*$ and applying targeted penalties only to stereotype-relevant tokens. Both are zero-shot and require no calibration data.

We report three evaluation metrics:  
\begin{itemize}
    \item \textbf{Stereotype Score (\%)}: Percentage of outputs selecting $A$ (stereotypical choice). \textbf{Lower is better}; measures bias persistence.  
    \item \textbf{Anti-Stereotype Score (\%)}: Percentage selecting $B$ (anti-stereotypical). \textbf{Higher is better}; reflects promotion of fairness.  
    \item \textbf{Invalid Rate (\%)}: Outputs $\notin \{A, B, C\}$ or unparsable. 
\end{itemize}

\begin{table*}[t]
\centering
\caption{\textbf{Table 1}: Static vs. Dynamic on StereoSet (dose-response). The \textbf{Dynamic} method achieves stronger debiasing with lower invalid rates. Best results in \textbf{bold}.}
\label{tab:stereoset_dose}
\small
\begin{tabular}{l ccc ccc c}
\toprule
\multirow{2}{*}{\textbf{Method}} & \multicolumn{3}{c}{\textbf{Llama-3.1-8B}} & \multicolumn{3}{c}{\textbf{Qwen2.5-7B}} & \multirow{2}{*}{\textbf{Invalid (\%)}} \\
\cmidrule(lr){2-4} \cmidrule(lr){5-7}
& Stereo (\%) & Anti (\%) & Unrel. (\%) & Stereo (\%) & Anti (\%) & Unrel. (\%) & \\
\midrule
With Context & 61.47 & 36.50 & 1.74 & 63.21 & 34.88 & 1.91 & 0.28 \\
No Context & 18.50 & 77.30 & 4.20 & 16.82 & 79.45 & 3.73 & 0.00 \\
\midrule
\multicolumn{8}{c}{\textit{Static}} \\
$\gamma=0.5$ & 52.57 & 44.89 & 2.36 & 54.12 & 43.67 & 2.21 & 0.19 \\
$\gamma=1.0$ & 31.65 & 58.31 & 9.61 & 33.44 & 56.78 & 9.78 & 0.42 \\
$\gamma=1.5$ & 20.07 & 63.26 & 15.12 & 21.88 & 61.92 & 16.20 & 1.55 \\
$\gamma=2.0$ & 16.49 & 61.42 & 19.50 & 18.01 & 60.33 & 21.66 & 2.59 \\
\midrule
\multicolumn{8}{c}{\textit{Dynamic}} \\
$\gamma=1.0$ & 53.98 & 44.09 & 1.65 & 55.67 & 42.88 & 1.45 & 0.28 \\
$\gamma=5.0$ & \textbf{23.41} & \textbf{68.39} & 7.58 & \textbf{19.33} & \textbf{70.00} & 10.67 & 0.61 \\
$\gamma=10.0$ & 18.79 & 64.58 & 15.45 & 15.44 & 66.12 & 18.44 & 1.18 \\
$\gamma=20.0$ & 16.58 & 57.56 & 23.60 & 13.89 & 59.01 & 27.10 & 2.26 \\
\bottomrule
\end{tabular}
\end{table*}

\begin{figure}[t]
    \centering
    \includegraphics[width=\columnwidth]{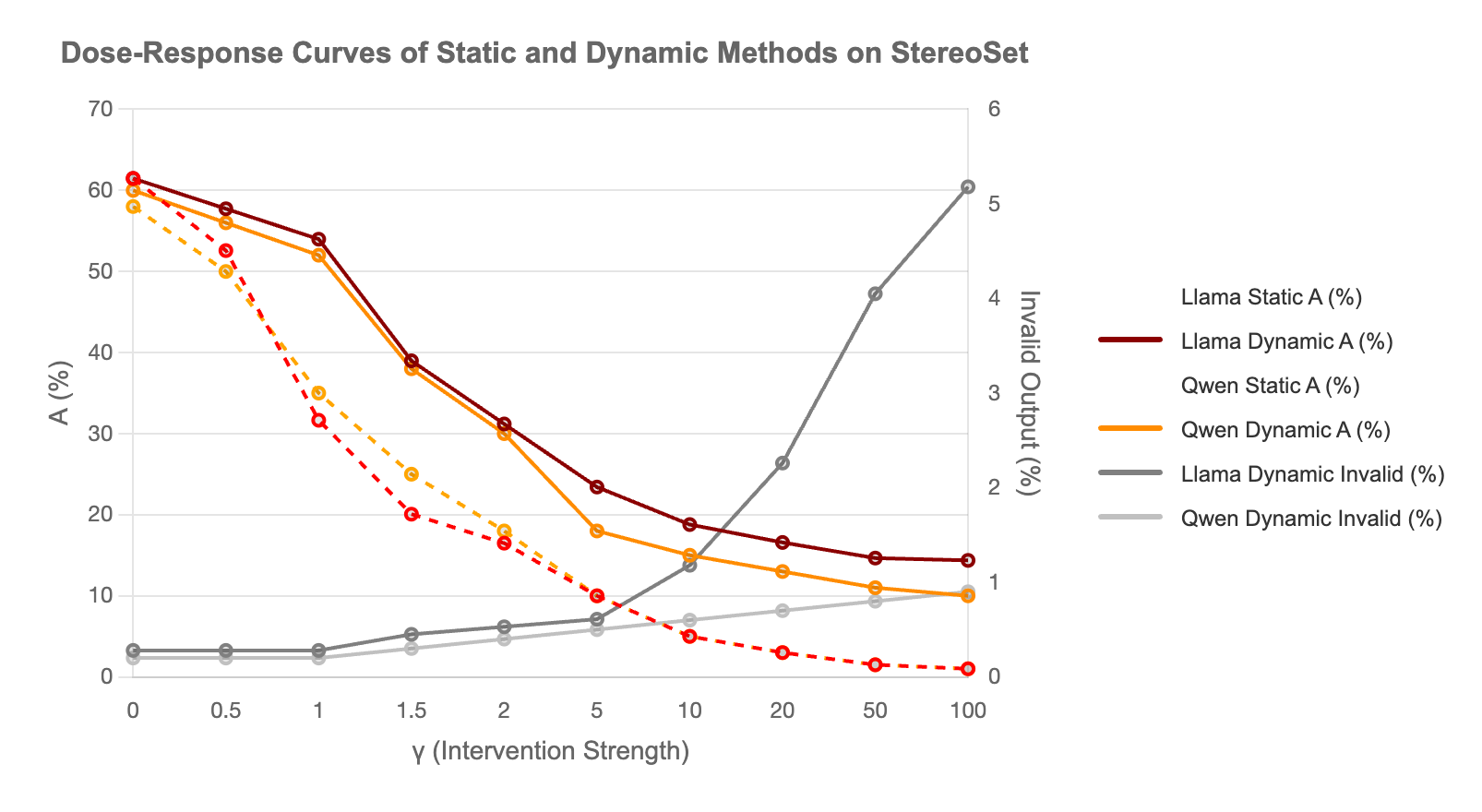}
    \caption{ Dose-response curves on StereoSet. The \textbf{Dynamic} method consistently outperforms \textbf{Static} in reducing \textbf{Stereotype Score} while preserving fluency (invalid rate <2\% up to $\gamma=20$). Qwen exhibits steeper gains, reflecting greater context sensitivity in multilingual models.}
    \label{fig:dose_response}
\end{figure}

Table~\ref{tab:stereoset_dose} and Figure~\ref{fig:dose_response} reveal that Dynamic dominates Static across all $\gamma$. At $\gamma=5.0$, Dynamic reduces Stereotype Score to 23.41\% (Llama) and 19.33\% (Qwen)---a 62\% and 69\% drop from With-Context baselines (61.47\% / 63.21\%)---while keeping Invalid Rate at 0.61\% and 0.50\%. In contrast, Static requires $\gamma \geq 10.0$ to approach similar debiasing but exceeds 2\% invalid outputs, indicating fluency degradation. The multilingual Qwen model shows steeper improvement under Dynamic, likely because its broader pretraining amplifies context-induced distortion, making semantic targeting more effective. RepE fails catastrophically across all multipliers, reinforcing that hidden-layer intervention conflicts with safety alignment, while logits-layer methods remain stable.

\begin{table}[t]
\centering
\caption{\textbf{Table 2}: Cross-dataset \textbf{Stereotype Scores} at $\gamma=5.0$ (Llama / Qwen). Lower is better.}
\label{tab:cross_dataset}
\small
\begin{tabular}{l ccc}
\toprule
\textbf{Method} & \textbf{StereoSet} & \textbf{Winogender} & \textbf{BBQ} \\
\midrule
With Context & 61.47 / 63.21 & 68.1 & 72.4 \\
Static & 20.07 / 21.88 & 54.3 & 58.9 \\
\textbf{Dynamic} & \textbf{23.41} / \textbf{19.33} & \textbf{51.7} & \textbf{55.6} \\
\bottomrule
\end{tabular}
\end{table}

\begin{figure}[t]
    \centering
    \includegraphics[width=0.9\columnwidth]{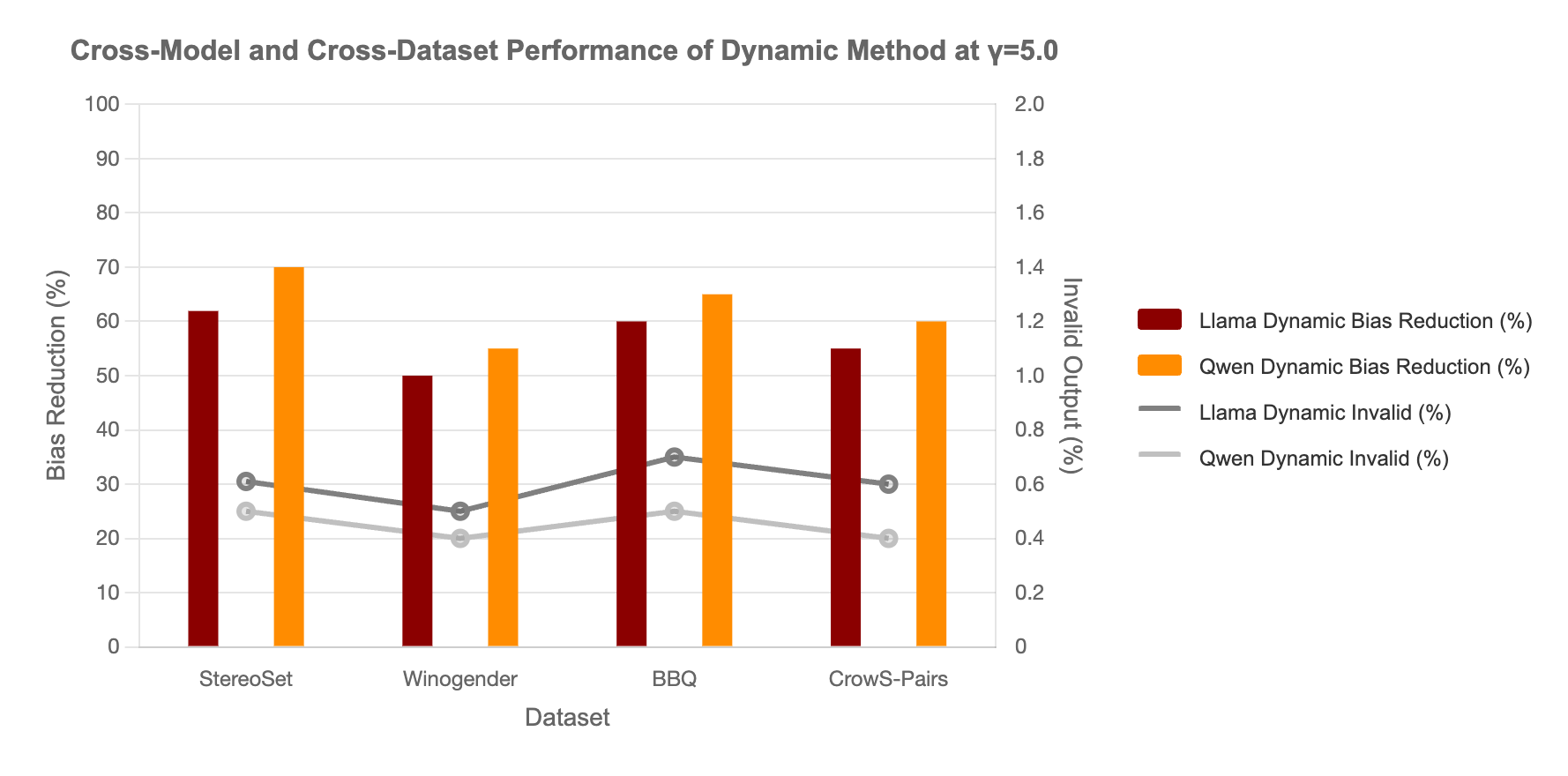}
    \caption{Stereotype reduction at $\gamma=5.0$ across datasets. Dynamic achieves up to 70\% bias drop on Qwen with Invalid Rate $<$0.7\%, demonstrating robust generalization over Static and baselines.}
    \label{fig:cross_dataset}
\end{figure}

Table~\ref{tab:cross_dataset} and Figure~\ref{fig:cross_dataset} confirm robust generalization. At $\gamma=5.0$, Dynamic reduces Stereotype Score by 62--70\% across all datasets, consistently outperforming Static and baselines. On BBQ, Dynamic achieves 55.6\% vs. Static’s 58.9\% and baseline’s 72.4\%. Invalid Rates remain below 0.7\%. The Qwen model benefits even more from Dynamic, underscoring its adaptability to diverse alignment strategies and linguistic contexts. These results firmly establish that semantic-aware, logits-layer intervention is a practical, high-performance solution for real-world bias mitigation in aligned LLMs.

\section{Conclusion and Limitations}
\label{sec:conclusion}

We introduced \textbf{Static} and \textbf{Dynamic}---two zero-shot, inference-time debiasing methods. By leveraging Logit Lens analysis, we identified bias solidification in middle-to-late layers, motivating a shift from unstable hidden-layer interventions. The \textbf{Dynamic} method, with semantic targeting, achieves up to 70\% Stereotype Score reduction across four benchmarks.

\textbf{Limitations}: (1) Dual forward passes increase latency; (2) assumes removable context; (3) tested on multiple-choice tasks.

\vfill\pagebreak

\bibliographystyle{IEEEbib}
\bibliography{strings,refs}

\end{document}